\documentclass[10pt,journal,compsoc]{IEEEtran-customize}

\ifCLASSOPTIONcompsoc
  \usepackage[nocompress]{cite}
\else
  \usepackage{cite}
\fi
\usepackage{graphicx}
\usepackage{color}
\usepackage{xcolor}
\usepackage{booktabs}
\usepackage{multirow}
\usepackage[hyphens]{url}
\usepackage{hyperref}
\usepackage{caption}
\usepackage{amsmath}
\usepackage{subcaption}
\usepackage{amsfonts, amssymb}
\usepackage{colortbl}
\usepackage{enumitem}

\definecolor{purple}{RGB}{128, 0, 128}
\definecolor{LightRed}{rgb}{1,0.92,0.92}
\definecolor{LightOrange}{rgb}{1,0.95,0.88}
\definecolor{LightYellow}{rgb}{1.0,1.0,0.84}
\definecolor{LightGreen}{rgb}{0.9,1.0,0.88}
\definecolor{LightCyan}{rgb}{0.9,1,1}
\definecolor{LightBlue}{rgb}{0.9,0.94,1}
\definecolor{LightIndigo}{rgb}{0.92,0.9,1}
\definecolor{LightMagenta}{rgb}{0.96,0.86,1}
\definecolor{DirtyWhite}{rgb}{0.96,0.96,0.96}

\DeclareSymbolFont{extraup}{U}{zavm}{m}{n}
\DeclareMathSymbol{\varheart}{\mathalpha}{extraup}{86}
\DeclareMathSymbol{\vardiamond}{\mathalpha}{extraup}{87}
\DeclareMathSymbol{\varclubsuit}{\mathalpha}{extraup}{88}

\begin{document}

\title{Computed Tomography Visual Question Answering with Cross-modal Feature Graphing}

\author{
Yuanhe Tian$^{\varheart}$, \hspace{0.1cm}
  Chen Su$^{{\spadesuit}}$, \hspace{0.1cm}
    Junwen Duan$^{\lozenge}$, \hspace{0.1cm}
    Yan Song$^{{\spadesuit}*}$\thanks{$^*$ Corresponding Author}
    \\
    $^{\varheart}$University of Washington
    \hspace{0.1cm}
    $^{\spadesuit}$University of Science and Technology of China \\
    $^{\lozenge}$Central South University \\
    $^{\varheart}$\texttt{yhtian@uw.edu} \hspace{0.1cm}
    $^{\spadesuit}$\texttt{suchen4565@mail.ustc.edu.cn}\\
    $^{\lozenge}$\texttt{jwduan@csu.edu.cn}
    \hspace{0.1cm}
    $^{\spadesuit}$\texttt{clksong@gmail.com}
}

\IEEEtitleabstractindextext{%

\begin{abstract}

Visual question answering (VQA) in medical imaging aims to support clinical diagnosis by automatically interpreting complex imaging data in response to natural language queries. 
Existing studies typically rely on distinct visual and textual encoders to independently extract features from medical images and clinical questions, which are subsequently combined to generate answers. 
Specifically, in computed tomography (CT), such approaches are similar to the conventional practices in medical image analysis. 
However, these approaches pay less attention to the spatial continuity and inter-slice correlations in the volumetric CT data, leading to fragmented and imprecise responses.
In this paper, we propose a novel large language model (LLM)-based framework enhanced by a graph representation of salient features. 
Different from conventional multimodal encoding strategies, our approach constructs a cross-modal graph integrating both visual and textual features, treating individual CT slices and question tokens as nodes within the graph. 
We further leverage an attentive graph convolutional network to dynamically fuse information within this structure. 
The resulting aggregated graph features then serve as a soft prompt to guide a large language model in generating accurate answers. 
Extensive experiments on the M3D-VQA benchmark demonstrate that our approach consistently outperforms baselines across multiple evaluation metrics, offering more robust reasoning capabilities.
\end{abstract}

\thanks{The code is available at \url{https://github.com/synlp/A-GCN-CTVQA}.}

\begin{IEEEkeywords}
Visual question answering, medical question answering, large language models, graph modeling
\end{IEEEkeywords}
}

\maketitle
\IEEEdisplaynontitleabstractindextext
\IEEEpeerreviewmaketitle

\makeatletter
\def\@IEEEcompsocmakefnmark{\hbox{\normalfont\@thefnmark\ }}
\long\def\@makefntext#1{\parindent 1em\indent\hbox{\@IEEEcompsocmakefnmark}#1}
\makeatother

\makeatletter
\def\@IEEEcompsocmakefnmark{\hbox{\normalfont\@thefnmark.\ }}
\long\def\@makefntext#1{\parindent 1em\indent\hbox{\@IEEEcompsocmakefnmark}#1}
\makeatother

\renewcommand{\thefootnote}{\arabic{footnote}}

\section{Introduction}

\begin{figure}
    \centering
    \includegraphics[width=1\linewidth, trim=0 20 0 0]{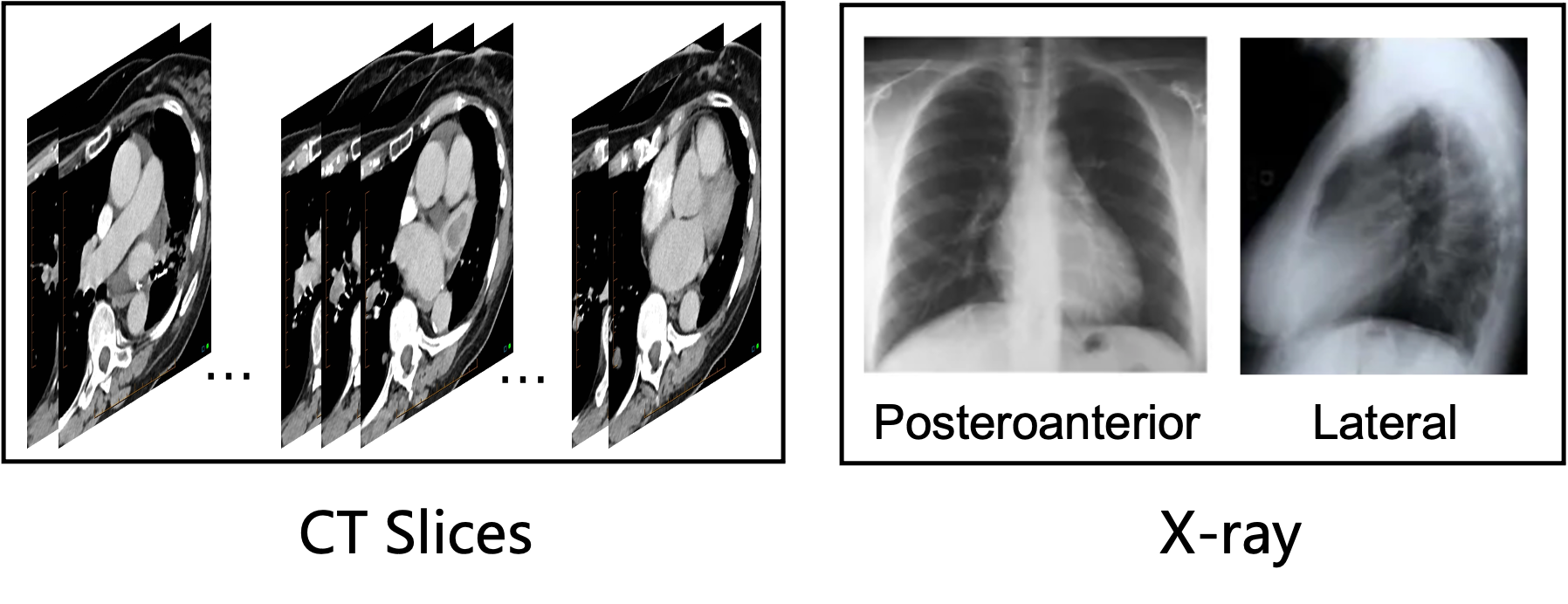}
    \caption{
    Comparison between CT slices and X-ray radiographs, where a CT volume generally contains much more images than an X-ray radiograph case.
    }
    \label{fig:intro-example}
    \vspace{-1.5em}
\end{figure}

Visual question answering (VQA) attracts significant attention in both academic and industrial communities, owing to its capacity to integrate computer vision and natural language processing for knowledge inference. 
Particularly, in the context of medical imaging, VQA holds immense potential for assisting clinical diagnosis and decision-making.
Especially for computed tomography (CT), its multi-slice nature offers VQA greater value since processing many images at a time costs physicians much labor and often causes problems such as inaccurate interpretation of the input or false answers.
As illustrated in Figure \ref{fig:intro-example}, CT scans provide three-dimensional information by stacking multiple slices, providing rich anatomical and pathological details, which significantly differs from typical conventional images (e.g., X-ray radiographs).
Therefore, CT VQA requires accurate analysis of multiple CT slices rather than one or a few images,
so that raises substantial challenges that the VQA model needs to handle voluminous data, effectively leverage cross-slice correlations, and interpret the spatial complexity across slices to accurately answer clinically relevant questions.

\begin{figure*}[t]
    \centering
    \includegraphics[width=0.98\textwidth, trim=0 10 0 0]{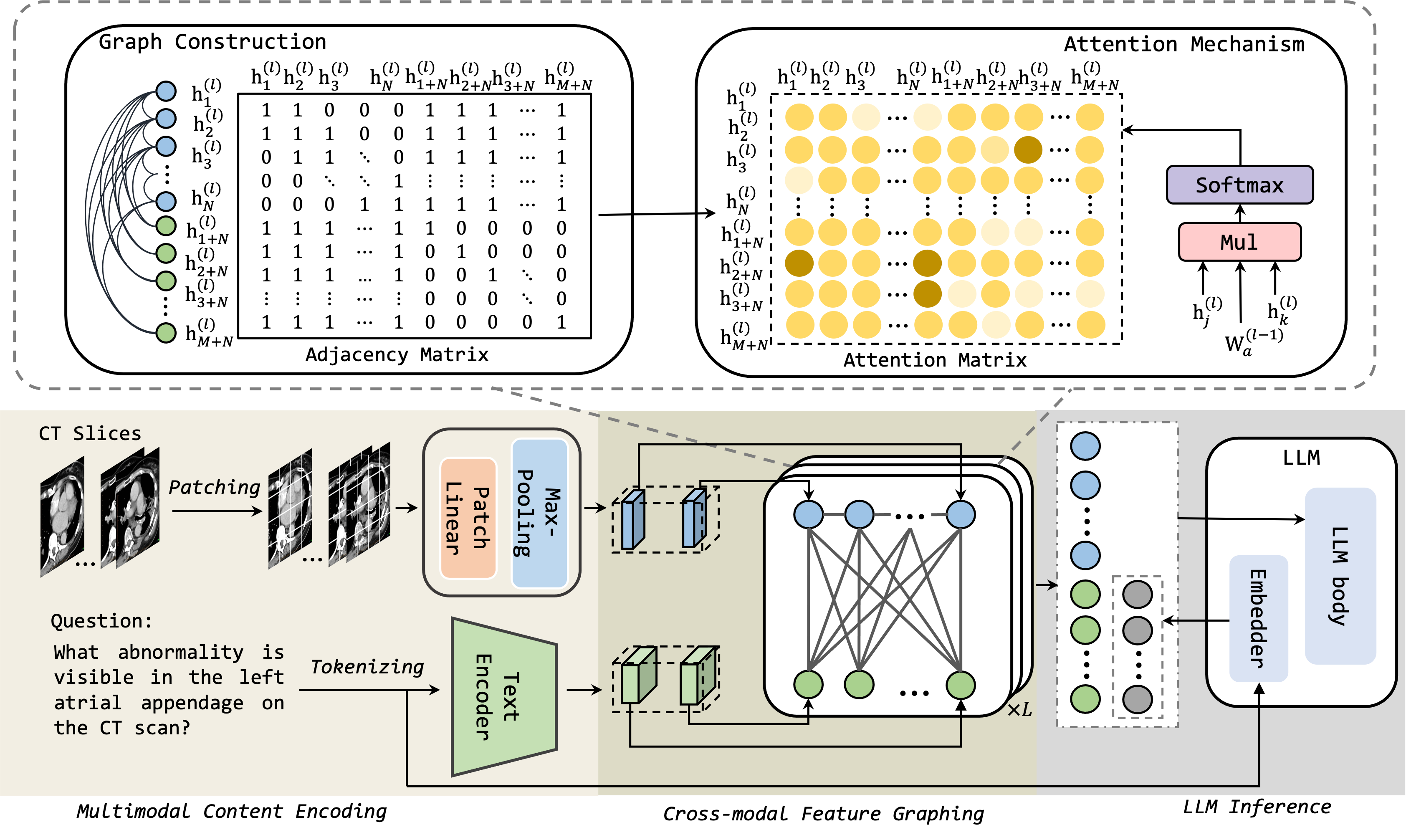}
    \caption{\textcolor{black}{
    An overview of our approach for CT VQA.
    The bottom part illustrates the input CT slices with the question, the vision and text encoder, the cross-modal feature graphing, and the LLM to generate the answer.
    The top part presents the graph built over the CT slices and question tokens, as well as the attention mechanism used in the cross-modal feature graphing for distinguishing important slices for answering the question.
    }}
    \label{fig:pipeline}
    \vspace{-1.5em}
\end{figure*}

Existing medical VQA approaches largely build upon general-domain versatile architectures by combining visual and textual encoders.
In doing so, a deep vision encoder extracts features from medical images, while a text encoder processes the clinical query; these features are then fused via attention mechanisms or joint embedding strategies to generate answers \cite{Antol2015,Abacha2019}. 
Based on the conventional structures, researchers further improve medical VQA by leveraging transfer learning, multi-modal pretraining, and meta-learning, as well as incorporating external domain knowledge through medical ontologies and knowledge graphs \cite{Liu2018,AlSadi2021,Lin2023}.
While these studies achieve impressive results on normal medical images, directly applying them to CT volumes introduces significant challenges owing to the fact that
a CT scan is composed of multiple adjacent image slices.
Therefore, conventional approaches are limited in their model structure to treat the volume as a set of unrelated 2D images, so that fail to capture the continuity and inherent spatial relationships among consecutive slices \cite{Lee2023}.
This lack of integration across slices weakens the model’s ability to perform robust spatial and quantitative reasoning, which causes the generated answers to be clinically superficial or inaccurate \cite{Nan2023}.
Moreover, existing studies pay less attention to identifying the important slices that are relevant to the key content in the question, which makes it hard for the model to locate essential information that is relevant to the question.

In this paper, we propose an LLM-based approach for CT VQA with cross-modal feature graphing.
Specifically, we utilize a vision encoder and a text encoder to capture the visual and textual representations of the CT slices and the text tokens,
then build a graph based on the representations using the information across modalities with each slice or token serving as a node.
To facilitate graph encoding,
we not only model the connections between every two nodes but also weight them accordingly to assess their contributions in predicting the correct answer, especially the images associated with the key nodes in the query.
This design allows the model to effectively identify CT slices that are relevant to the question, reducing noise from irrelevant slices.
Finally, the encoded representation from the graph of CT slices and key information in the question is passed to an LLM for answer generation.
Extensive experiments on a benchmark CT VQA dataset, namely, M3D-VQA, demonstrate the effectiveness of our approach, which outperforms strong baselines and existing studies.

\section{The Approach}

Our approach for CT VQA enhances multimodal LLM by cross-modal feature graphing, which is modeled by graph convolutional networks (GCN), with the overall architecture presented in Figure \ref{fig:pipeline}.
Specifically, given the input CT volume $\mathcal{V} = \mathcal{S}_1 \cdots \mathcal{S}_N$ that is the combination of $N$ CT slices (the $n$-th CT slice is denoted as $\mathcal{S}_n$) and the question $\mathcal{Q} = q_1 \cdots q_M$ with $M$ tokens (the $m$-th token is denoted as $q_m$), 
our approach firstly perform a multimodal content encoding process ($f_{e}$) to compute their embeddings.
Next, the graph is constructed, where each node represents either CT slices or question tokens, 
and the representation of the graph structure is derived from the embedding of CT slices and question tokens.
Subsequently, we apply an attentive graph convolutional network (A-GCN, denoted as $f_{AGCN}$) to encode the graph, weighing different nodes to identify the most relevant information for the QA task.
Finally, the output of the graph structure decoding is used as a soft prompt to instruct the LLM ($f_{LLM}$) in generating the answer $\widehat{\mathcal{A}}$.
Therefore, the overall process of our approach is formulated as
\begin{equation}
    \widehat{\mathcal{A}} = f_{LLM} (f_{AGCN} (f_e(\mathcal{V}, \mathcal{Q})), \mathcal{Q})
\end{equation}
In the following text, we firstly present the multimodal content encoding process, then the cross-modal feature graph modeling, and finally the LLM decoding process for answer generation.

\subsection{Multimodal Content Encoding}

The multimodal content encoding process aims to process the CT slices $\mathcal{S}_1 \cdots \mathcal{S}_N$ and the question $\mathcal{Q}$ into their representations so as to facilitate the following processing.
In doing so, we utilize a vision encoder $f_v$ (e.g., a pre-trained vision Transformer \cite{chen2021vit}) and a text encoder $f_t$ (e.g., BERT \cite{devlin-etal-2019-bert}) to model the CT slices and question, respectively.
\textcolor{black}{
For the vision encoding process, for each CT slice $\mathcal{S}_n$, we firstly split it into non-overlapping patches $p_{n,1} \cdots p_{n, U}$ and use a linear projection function to map the pixel value of each patch into its patch embedding, where the $u$-th one is denoted as $\mathbf{e}^p_{n,u}$.
Then, we stack all patch embeddings and use the vision Transformer to process them by
\begin{equation}
    \mathbf{H}^s_n = f_v (\mathbf{e}^p_{n,1} \cdots \mathbf{e}^p_{n,U})
\end{equation}
where $\mathbf{H}^s_n$ is the output matrix from the last layer of the vision Transformer, where the column vector stands for the encoded representation for the corresponding patch.
We perform a max-pooling operation on $\mathbf{H}^s_n$ and obtain a vector $\mathbf{h}^s_n$ that represents the $n$-th slice by
\begin{equation}
    \mathbf{h}^s_n = \texttt{Max-Pooling} (\mathbf{H}^s_n)
\end{equation}
We perform the aforementioned process for all CT slices and obtain their representations $\mathbf{h}^s_1 \cdots \mathbf{h}^s_N$.
}

\textcolor{black}{
For the question $\mathcal{Q} = q_1 \cdots q_M$, we utilize an embedding layer to map the tokens $q_m$ in $\mathcal{Q}$ into their embeddings, which is denoted as $\mathbf{e}^q_m$ for the $m$-th token.
Then, we stack the embeddings of all tokens and pass them through the text encoder by
\begin{equation}
    \mathbf{H}^q = f_t (\mathbf{e}^q_1 \cdots \mathbf{e}^q_M)
\end{equation}
where $\mathbf{H}^q = \mathbf{h}^q_1 \cdots \mathbf{h}^q_M$ with the representation of the $m$-th token denoted as $\mathbf{h}^q_m$.
The CT slice representations (i.e., $\mathbf{h}^s_1 \cdots \mathbf{h}^s_N$) and the question token representations (i.e., $\mathbf{h}^q_1 \cdots \mathbf{h}^q_M$) are used in the cross-modal feature graph modeling.
}

\subsection{Cross-modal Feature Graphing}

\textcolor{black}{
For medical CT visual question answering (VQA), different questions may focus on different anatomical regions in a CT scan, and different CT slices capture distinct imaging characteristics of these regions.
Therefore, it is crucial to assess the relevance of CT slices as well as their relations to the question, particularly in relation to key terms in the question.
This helps in selecting the most relevant CT slices and utilizing them to improve the answering process.
Consider graph-based models are demonstrated to be effective in modeling the relations \cite{zhang2020radiology,tian-etal-2020-supertagging,qin-etal-2021-relation,li2023dynamic}, 
we construct a graph structure connecting CT slices and question tokens and propose a cross-modal feature graphing model. 
This approach firstly builds a graph over the CT slices and question tokens, and then employs convolutional graph networks to encode the graph, which is enhanced by an attention mechanism to distinguish important information that is essential for answering the question from the unimportant ones.
The details are elaborated as follows.
}

\subsubsection{Graph Construction}
\textcolor{black}{
For graph construction,
we represent CT slices and question tokens as nodes in the graph, with their representations denoted as $\mathbf{h}_1, \cdots, \mathbf{h}_{M+N}$.
For $j \leq N$, the representation of the $j$-th node corresponds to the representation of the $j$-th CT slice;
for $N < j \leq M+N$, the representation of the $j$-th node corresponds to the representation of the $(j-N)$-th token in the question.
This is formulated as
\begin{equation}
\mathbf{h}_j =
\begin{cases}
\mathbf{h}^s_j, & \text{if } j \leq N \\
\mathbf{h}^q_{j-N}, & \text{if } N < j \leq M+N
\end{cases}
\end{equation}
Next, we establish edges between the nodes.
Since adjacent CT slices often share strong spatial and structural relationships, we connect neighboring CT slices with edges.
Additionally, as question tokens can relate to multiple CT slices, we establish edges between every token and every slice.
We define an adjacency matrix $\mathbf{A}$ to represent the connections between nodes, where the entry at position $(j, k)$, denoted as $a_{j,k}$, is either 0 or 1.
A value of 0 indicates no edge between the $j$-th and $k$-th nodes, while a value of 1 denotes the presence of an edge.
}

\begin{table}[t]
\centering
\caption{
\textcolor{black}{
The statistics of the M3D-VQA dataset used in our experiments, where the total number of CT volume, the CT Slices, and questions, as well as the average number of slices and questions per CT volume, are reported.}
}
\begin{tabular}{l|ccc}
\toprule
 & \textbf{Train} & \textbf{Dev} & \textbf{Test} \\
\midrule
\# of CT Volume
& 92,170 & 2,000 & 2,000 \\
\# of CT Slices 
& 8,042,725 & 224,891 & 171,556 \\
\# of Questions 
& 481,897 & 14,067 & 13,791 \\
\midrule
Avg. \# of Slices per CT Volume
& 69.3 & 122.4 & 85.8  \\
Avg. Questions per CT Volume
& 5.2 & 7.0 & 6.9 \\
\bottomrule
\end{tabular}
\label{tab:dataset}
\end{table}

\begin{table*}[t]
  \centering
      \caption{\textcolor{black}{
    The overall results (i.e., BLEU, ROUGE, METEOR, and BERT-Score) of different baselines and our approach (with Qwen2-VL (2B)) on various question types, where the average performance on all question types is also reported.
    }}
    \label{tab:overall}
  \begin{tabular}{l | c | c c c c c | c}
    \toprule
    Approach & Metric & Plane & Phase & Organ & Abnormality & Location & Mean \\
    \midrule
    \multirow{4}{*}{LLM}
      & BLEU       & 97.87 & 74.02 & 33.82 & 15.46 & 23.26 & 48.89 \\
      & ROUGE      & 97.90 & 78.25 & 37.59 & 18.79 & 27.30 & 51.97 \\
      & METEOR     & 48.73 & 63.41 & 23.30 & 12.45 & 18.09 & 33.20 \\
      & BERT-Score & 98.94 & 95.16 & 88.53 & 85.71 & 87.13 & 91.09 \\
    \midrule
    \multirow{4}{*}{LLM + GCN}
      & BLEU       & 98.03 & 74.28 & 34.02 & 15.68 & 23.64 & 49.13 \\
      & ROUGE      & 97.99 & 78.42 & 37.71 & 19.02 & 27.49 & 52.13 \\
      & METEOR     & 48.96 & 63.50 & 23.45 & 12.58 & 18.27 & 33.35 \\
      & BERT-Score & 99.10 & 95.31 & 88.77 & 85.93 & 87.29 & 91.28 \\
    \midrule
    \multirow{4}{*}{LLM + GAT}
      & BLEU       & 98.25 & 74.48 & 34.21 & 15.83 & 23.91 & 49.34 \\
      & ROUGE      & 98.23 & 78.51 & 37.95 & 19.17 & 27.57 & 52.29 \\
      & METEOR     & 49.11 & 63.56 & 23.63 & 12.65 & 18.53 & 33.50 \\
      & BERT-Score & 99.27 & 95.40 & 89.02 & 86.17 & 87.41 & 91.45 \\
    \midrule
    \multirow{4}{*}{Ours} 
      & BLEU       & \textbf{98.57} & \textbf{74.72} & \textbf{34.58} & \textbf{16.33} & \textbf{24.43} & \textbf{49.73} \\
      & ROUGE      & \textbf{98.52} & \textbf{78.81} & \textbf{38.39} & \textbf{19.52} & \textbf{28.02} & \textbf{52.65} \\
      & METEOR     & \textbf{49.51} & \textbf{63.91} & \textbf{24.03} & \textbf{13.18} & \textbf{18.84} & \textbf{33.89} \\
      & BERT-Score & \textbf{99.62} & \textbf{95.73} & \textbf{89.45} & \textbf{86.52} & \textbf{87.72} & \textbf{91.81} \\
    \bottomrule
  \end{tabular}
\end{table*}

\subsubsection{Graph Encoding}
In graph encoding, the standard approach utilizes GCN, whose
input consists of the representations of all nodes $\mathbf{h}_1, \cdots, \mathbf{h}_{M+N}$ and the adjacency matrix $\mathbf{A}$, and the output is the hidden vectors of these nodes computed at the final layer, denoted as $\mathbf{h}^{(L)}_1, \cdots, \mathbf{h}^{(L)}_{M+N}$.
The A-GCN consists of $L$ layers, where the input to each layer is the output from the previous layer.
Specifically, for the $l$-th layer and the $j$-th node, the output $\mathbf{h}^{(l)}_j$ is computed as:
\begin{equation} \label{eq:gcn}
    \mathbf{h}^{(l)}_j = \sum_{k=1}^{M+N} a_{j,k} \cdot \mathbf{W}^{(l)} \cdot \mathbf{h}^{(l-1)}_k + \mathbf{b}^{(l)}
\end{equation}
where $\mathbf{W}^{(l)}$ and $\mathbf{b}^{(l)}$ are the trainable weighting matrix and the bias vector for the $l$-th layer, and $\mathbf{h}^{(l-1)}_k$ represents the input from node $k$ at the $(l-1)$-th layer.

\subsubsection{Node Attention}
\textcolor{black}{
For CT VQA, the question generally focuses on a particular aspect of the CT volume, so that not all slices contribute equally to answering the question.
Therefore, we introduce an attention mechanism to the standard GCN, where an attention weight $w_{j,k}$ is used to replace $a_{j,k}$ in Eq. (\ref{eq:gcn}).
The term $w_{j,k}$ represents the attention weight that quantifies the relationship between the $j$-th and $k$-th nodes,
which is computed as
\begin{equation}
    w_{j,k} = \frac{a_{j,k} \cdot \exp (\mathbf{h}^{(l-1)}_j \cdot \mathbf{W}^{(l)}_{a} \cdot \mathbf{h}^{(l-1)}_k)}
    {\sum_{k=1}^{M+N}(\exp (\mathbf{h}^{(l-1)}_j \cdot \mathbf{W}^{(l)}_{a} \cdot \mathbf{h}^{(l-1)}_k))}
\end{equation}
where $\mathbf{W}^{(l)}_{a}$ is a trainable weight matrix used to compute attention scores.
Through this approach, we assign weighted importance to the relationships between different CT slices and question tokens, allowing the model to identify the most relevant slice-token associations.
Finally, we apply this computation iteratively across all layers until obtaining the last-layer outputs $\mathbf{h}^{(L)}_1, \cdots, \mathbf{h}^{(L)}_{M+N}$,
which serves as a prompt for the LLM for answer generation.
}

\subsection{Large Language Model Inference}

Upon the output of A-GCN, 
LLM inference produces the answer $\widehat{\mathcal{A}}$ for the question $\mathcal{Q}$.
In doing so, we firstly stack all $\mathbf{h}^{(L)}_1, \cdots, \mathbf{h}^{(L)}_{M+N}$ to construct a representation matrix $\mathbf{H}^{L} = \mathbf{h}^{(L)}_1, \cdots, \mathbf{h}^{(L)}_{M+N}$ and utilize a fully connected layer to map it to the embedding space of the LLM through
\begin{equation}
    \mathbf{O} = \mathbf{W} \cdot \mathbf{H^{(L)}} + \mathbf{b}
\end{equation}
where $\mathbf{O}$ is the representation that contains information that is essential for producing the answer, and $\mathbf{W}$ and $\mathbf{b}$ are trainable parameters of the fully connected layer.
Then, we use the embedding layer of the LLM to map the question $\mathcal{Q}$ into the embedding matrix $\mathbf{E}$, where each column represents the embedding of the corresponding token.
We feed $\mathbf{O}$ and $\mathbf{E}$ to the LLM and generate the answer following the standard LLM decoding process by
\begin{equation}
    \widehat{\mathcal{A}} = f_{LLM} (\mathbf{O}, \mathbf{E})
\end{equation}
In training, we compare the generated answer $\widehat{\mathcal{A}}$ with the gold standard one $\mathcal{A}^*$ to compute the cross-entropy loss for parameter optimization.

\section{Experiment Settings}

\subsection{Datasets}

The M3D-VQA dataset \cite{bai2024m3d} is utilized to evaluate our approach, with its statistics summarized in Table \ref{tab:dataset}.
M3D-VQA is a large-scale benchmark designed explicitly for visual question answering on CT scans, where
each instance in the dataset consists of a volumetric CT image set paired with clinically relevant questions \textcolor{black}{with multi-choice and open-ended answers} derived from expert-annotated reports.
The questions cover critical aspects such as imaging planes, phases, organs, abnormalities, and lesion localization, demanding detailed and nuanced analyses of anatomical structures and pathological findings.
Moreover, the question design in the M3D-VQA dataset captures the complexity of real clinical scenarios, challenging models not only on their comprehension of CT volume but also on their capacity for clinical reasoning.
We follow existing studies \cite{bai2024m3d} to split the dataset into train/dev/test and utilize the open-ended answer as the target answer. 
The statistics of the dataset are presented in Table \ref{tab:dataset}.

\subsection{Baselines}

In our experiments, we compare the proposed LLM + A-GCN approach with the following baselines:
\begin{itemize}
    \item \textbf{LLM:} This baseline performs CT VQA by directly using a multimodal large language model to encode CT slices and questions with standard visual and text encoders, and then generating answers without any other help.
    \item \textbf{LLM + GCN:} This variant employs standard graph convolutional networks (GCN) \cite{kipf2016semi} to model the relationships among CT slices and question tokens.
    \item \textbf{LLM + GAT:} This baseline incorporates graph attention networks (GAT) \cite{velivckovic2017graph} to dynamically weigh the contributions of different nodes, thus allowing the model to focus on more relevant CT slices or question tokens.
\end{itemize}
Compared to the baselines, our proposed approach, \textbf{LLM + A-GCN}, integrates attentive graph convolutional networks that not only model the structured cross-slice relationships but also leverage attention to guide the aggregation process for more clinically meaningful VQA results.
Note that the LLM has multiple options in the experiments, which are introduced in Section \ref{sec: implementation}.

\begin{table*}[t]
  \centering
      \caption{\textcolor{black}{
    Comparison of our approach with existing studies on different types of questions on M3D-VQA data.
    }}
      \label{tab:sota}
  \begin{tabular}{l | c | c c c c c | c}
    \toprule
    Approach & Metric & Plane & Phase & Organ & Abnormality & Location & Mean \\
    \midrule
    \multirow{4}{*}{RadFM \cite{wu2023towards}} 
      & BLEU       & 14.24 & 14.25 & 14.24 & 15.64 & 23.58 & 16.39 \\
      & ROUGE      & 25.40 & 25.41 & 25.38 & 25.38 & 29.09 & 26.13 \\
      & METEOR     & 20.62 & 20.63 & 20.61 & 20.60 & 24.19 & 21.33 \\
      & BERT-Score & 92.68 & 92.04 & 86.79 & 85.84 & 86.26 & 88.72 \\
    \midrule
    \multirow{4}{*}{M3D \cite{bai2024m3d}} 
      & BLEU       & 98.37 & 74.41 & 34.20 & 15.91 & 24.00 & 49.38 \\
      & ROUGE      & 98.42 & 78.63 & 37.87 & 19.27 & 27.74 & 52.39 \\
      & METEOR     & 49.20 & 63.58 & 23.78 & 12.83 & 18.50 & 33.58 \\
      & BERT-Score & 99.47 & 95.55 & 88.97 & 86.08 & 87.60 & 91.53 \\
    \midrule
        \multirow{4}{*}{Ours} 
      & BLEU       & \textbf{98.57} & \textbf{74.72} & \textbf{34.58} & \textbf{16.33} & \textbf{24.43} & \textbf{49.73} \\
      & ROUGE      & \textbf{98.52} & \textbf{78.81} & \textbf{38.39} & \textbf{19.52} & \textbf{28.02} & \textbf{52.65} \\
      & METEOR     & \textbf{49.51} & \textbf{63.91} & \textbf{24.03} & \textbf{13.18} & \textbf{18.84} & \textbf{33.89} \\
      & BERT-Score & \textbf{99.62} & \textbf{95.73} & \textbf{89.45} & \textbf{86.52} & \textbf{87.72} & \textbf{91.81} \\
    \bottomrule
  \end{tabular}
    \vspace{-0.3cm}
\end{table*}

\subsection{Implementation Details}
\label{sec: implementation}

Pretrained models have demonstrated their effectiveness in modeling the data in different modalities \cite{mikolov2013efficient,song-etal-2017-learning,ijcai2018-607,su2019vl,song2021zen,wang-etal-2022-medclip,touvron2023llama,li2024llava,tian-etal-2024-dialogue,tian2024learning}.
Therefore, in our approach, we also use pre-trained models, namely, 
the Qwen2-VL (2B) model \cite{wang2024qwen2}, as the backbone for our approach. 
In our implementation, the vision encoder of Qwen2-VL is responsible for encoding CT slices; this encoder is constructed with 32 Transformer blocks and produces hidden representations of dimension 1,280. 
All CT slices are resized to \(224 \times 224\) to ensure compatibility with the encoder. 
For the language model component, we follow the original Qwen2-VL configuration, which comprises 28 Transformer layers and utilizes a hidden state size of 1,536. 
In order to evaluate the effect on the LLM scales and types, we further try two additional multi-modal LLMs, namely Qwen2-VL (7B) and LLaVA-med (7B), following the same paradigm to leverage the vision encoder and text decoder.
The vision encoder of the Qwen2-VL (7B) model consists of 32 Transformer layers with 3,584 hidden size, and the LLM decoder uses 28 Transformer layers configured with the same 3,584 hidden size.
The vision encoder of the LLaVA-med (7B) utilizes 24 layers of Transformer with 1,024 hidden size, and the LLM decoder (which is Mistral \cite{jiang2023mistral7b}) employs 32 layers of Transformer with 4,096 hidden size.
Herein, Qwen2-VL (7B) shares the same architecture with Qwen2-VL (2B) with more parameters; LLaVA-med (7B) utilizes a different architecture and is fine-tuned on the medical data.
Unless specified, the default LLM in the experiments is Qwen2-VL (2B).

\begin{table*}[htbp]
  \centering
  \caption{\textcolor{black}{%
    The results of baselines and our approach with various types of LLMs, namely, Qwen2-VL (7B) and LLaVA-med (7B).
  }}
  \label{tab:llm-types}
  \begin{tabular}{l | l | c | c c c c c | c}
    \toprule
    LLM & Approach & Metric & Plane & Phase & Organ & Abnormality & Location & Mean \\
    \midrule
    \multirow{16}{*}{Qwen2-VL (7B)}
    & \multirow{4}{*}{LLM}
      & BLEU       & 97.92 & 74.04 & 33.85 & 15.49 & 23.31 & 48.92 \\
      &              & ROUGE      & 97.91 & 78.29 & 37.63 & 18.82 & 27.34 & 52.00 \\
      &              & METEOR     & 48.77 & 63.42 & 23.32 & 12.47 & 18.15 & 33.23 \\
      &              & BERT-Score & 98.96 & 95.17 & 88.55 & 85.73 & 87.14 & 91.11 \\
    \cline{2-9}
    & \multirow{4}{*}{LLM + GCN}
      & BLEU       & 98.13 & 74.29 & 34.03 & 15.76 & 23.68 & 49.18 \\
      &              & ROUGE      & 98.02 & 78.45 & 37.77 & 19.06 & 27.52 & 52.16 \\
      &              & METEOR     & 49.01 & 63.53 & 23.49 & 12.62 & 18.31 & 33.39 \\
      &              & BERT-Score & 99.14 & 95.37 & 88.80 & 85.98 & 87.33 & 91.32 \\
    \cline{2-9}
    & \multirow{4}{*}{LLM + GAT}
      & BLEU       & 98.29 & 74.52 & 34.23 & 15.89 & 23.97 & 49.38 \\
      &              & ROUGE      & 98.31 & 78.56 & 37.97 & 19.22 & 27.62 & 52.34 \\
      &              & METEOR     & 49.21 & 63.62 & 23.68 & 12.69 & 18.55 & 33.55 \\
      &              & BERT-Score & 99.32 & 95.43 & 89.05 & 86.24 & 87.47 & 91.50 \\
    \cline{2-9}
    & \multirow{4}{*}{Ours}
      & BLEU       & \textbf{98.63} & \textbf{74.76} & \textbf{34.61} & \textbf{16.36} & \textbf{24.48} & \textbf{49.77} \\
      &              & ROUGE      & \textbf{98.58} & \textbf{78.88} & \textbf{38.42} & \textbf{19.58} & \textbf{28.05} & \textbf{52.70} \\
      &              & METEOR     & \textbf{49.56} & \textbf{63.98} & \textbf{24.06} & \textbf{13.24} & \textbf{18.82} & \textbf{33.93} \\
      &              & BERT-Score & \textbf{99.64} & \textbf{95.77} & \textbf{89.48} & \textbf{86.57} & \textbf{87.75} & \textbf{91.84} \\
    \midrule
    \multirow{16}{*}{LLaVA-med (7B)}
    & \multirow{4}{*}{LLM}
      & BLEU       & 97.94 & 74.07 & 33.87 & 15.53 & 23.34 & 48.95 \\
      &              & ROUGE      & 97.93 & 78.32 & 37.66 & 18.88 & 27.38 & 52.03 \\
      &              & METEOR     & 48.75 & 63.45 & 23.33 & 12.52 & 18.17 & 33.24 \\
      &              & BERT-Score & 98.92 & 95.19 & 88.53 & 85.71 & 87.17 & 91.10 \\
    \cline{2-9}
    & \multirow{4}{*}{LLM + GCN}
      & BLEU       & 98.15 & 74.27 & 34.04 & 15.74 & 23.72 & 49.18 \\
      &              & ROUGE      & 98.06 & 78.46 & 37.78 & 19.09 & 27.55 & 52.19 \\
      &              & METEOR     & 49.04 & 63.57 & 23.53 & 12.64 & 18.32 & 33.42 \\
      &              & BERT-Score & 99.15 & 95.41 & 88.82 & 85.95 & 87.35 & 91.34 \\
    \cline{2-9}
    & \multirow{4}{*}{LLM + GAT}
      & BLEU       & 98.35 & 74.55 & 34.28 & 15.81 & 23.95 & 49.39 \\
      &              & ROUGE      & 98.32 & 78.59 & 38.03 & 19.27 & 27.60 & 52.36 \\
      &              & METEOR     & 49.25 & 63.66 & 23.72 & 12.72 & 18.56 & 33.58 \\
      &              & BERT-Score & 99.32 & 95.46 & 89.08 & 86.29 & 87.46 & 91.52 \\
    \cline{2-9}
    & \multirow{4}{*}{Ours}
      & BLEU       & \textbf{98.65} & \textbf{74.78} & \textbf{34.63} & \textbf{16.38} & \textbf{24.49} & \textbf{49.79} \\
      &              & ROUGE      & \textbf{98.62} & \textbf{78.90} & \textbf{38.47} & \textbf{19.61} & \textbf{28.08} & \textbf{52.74} \\
      &              & METEOR     & \textbf{49.62} & \textbf{64.00} & \textbf{24.12} & \textbf{13.28} & \textbf{18.86} & \textbf{33.98} \\
      &              & BERT-Score & \textbf{99.65} & \textbf{95.79} & \textbf{89.52} & \textbf{86.60} & \textbf{87.78} & \textbf{91.87} \\
    \bottomrule
  \end{tabular}
\end{table*}

During training, we jointly update both the visual encoder and the language model in an end-to-end manner.
By default, we perform full parameter fine-tuning where all parameters in the model are updated.
We also try a representative parameter-efficient fine-tuning strategy named LoRA \cite{hu2021lora} in the experiments.
The model is trained for three epochs using a learning rate of \(5 \times 10^{-5}\), a batch size of 16, and a weight decay of 0.01. 
For answer generation, we always produce the token with the highest probability to ensure reproducible results. 
To evaluate the quality of generated answers, we use standard natural language generation metrics, including BLEU, ROUGE, METEOR, and BERT-score \cite{zhang2019bertscore}. 
While BLEU and ROUGE measure n-gram overlap with reference texts, METEOR accounts for stemming and synonymy, and BERT-score uses BERT-based model to compute token-level cosine similarities, thereby capturing semantic alignment. 
These metrics provide a comprehensive assessment of the answer quality.

\section{Results and Analysis}

\begin{table*}[htbp]
  \centering
      \caption{\textcolor{black}{
    The overall results of our approach with Qwen2-VL (2B) using full parameter fine-tuning and a representative effective fine-tuning approach named LoRA.
    }}
    \label{tab:peft}
  \begin{tabular}{l | l | c | c c c c c | c}
    \toprule
    Training Strategies & Approach & Metric & Plane & Phase & Organ & Abnormality & Location & Mean \\
    \midrule
    \multirow{8}{*}{Full Parameter Fine-tuning}
    & \multirow{4}{*}{LLM}
            & BLEU       & 97.87 & 74.02 & 33.82 & 15.46 & 23.26 & 48.89 \\
      && ROUGE      & 97.90 & 78.25 & 37.59 & 18.79 & 27.30 & 51.97 \\
      && METEOR     & 48.73 & 63.41 & 23.30 & 12.45 & 18.09 & 33.20 \\
      && BERT-Score & 98.94 & 95.16 & 88.53 & 85.71 & 87.13 & 91.09 \\
      \cline{2-9}
          & \multirow{4}{*}{Ours}
      & BLEU       & \textbf{98.57} & \textbf{74.72} & \textbf{34.58} & \textbf{16.33} & \textbf{24.43} & \textbf{49.73} \\
      && ROUGE      & \textbf{98.52} & \textbf{78.81} & \textbf{38.39} & \textbf{19.52} & \textbf{28.02} & \textbf{52.65} \\
      && METEOR     & \textbf{49.51} & \textbf{63.91} & \textbf{24.03} & \textbf{13.18} & \textbf{18.84} & \textbf{33.89} \\
      && BERT-Score & \textbf{99.62} & \textbf{95.73} & \textbf{89.45} & \textbf{86.52} & \textbf{87.72} & \textbf{91.81} \\
      \midrule
    \multirow{8}{*}{LoRA}
      & \multirow{4}{*}{LLM}
      & BLEU       & 97.83 & 73.97 & 33.79 & 15.42 & 23.24 & 48.85 \\
      && ROUGE      & 97.85 & 78.22 & 37.57 & 18.77 & 27.27 & 51.94 \\
      && METEOR     & 48.71 & 63.38 & 23.27 & 12.43 & 18.07 & 33.17 \\
      && BERT-Score & 98.92 & 95.12 & 88.51 & 85.68 & 87.11 & 91.07 \\
      \cline{2-9}
    & \multirow{4}{*}{Ours}
      & BLEU       & \textbf{98.54} & \textbf{74.70} & \textbf{34.55} & \textbf{16.31} & \textbf{24.40} & \textbf{49.70} \\
      && ROUGE      & \textbf{98.49} & \textbf{78.78} & \textbf{38.37} & \textbf{19.49} & \textbf{27.97} & \textbf{52.62} \\
      && METEOR     & \textbf{49.47} & \textbf{63.87} & \textbf{24.01} & \textbf{13.16} & \textbf{18.81} & \textbf{33.86} \\
      && BERT-Score & \textbf{99.58} & \textbf{95.71} & \textbf{89.42} & \textbf{86.50} & \textbf{87.68} & \textbf{91.78} \\
    \bottomrule
  \end{tabular}
\end{table*}

\subsection{Overall Results}

Table \ref{tab:overall} presents the overall performance of our proposed LLM + A-GCN model compared with the baselines across multiple evaluation metrics, including BLEU, ROUGE, METEOR, and BERT-Score. 
First, compared with the LLM baseline, our approach consistently achieves higher scores on all metrics.
This improvement indicates that incorporating explicit graph-based modeling of cross-slice and cross-modal relationships helps the model capture essential structural information that a vanilla LLM fails to exploit.
Second, our model also outperforms the LLM + GCN baseline. While the standard GCN aggregates information using a fixed neighborhood structure, our attentive mechanism within the A-GCN adaptively weights contributions from different nodes, leading to a more effective integration of relevant visual and textual features in CT VQA.
Third, compared with the LLM + GAT baseline, our approach shows superior performance. Although GAT employs dynamic attention to weight node contributions, it lacks the question related information in guidance,
which enables our full model to better focus on critical regions and clinical cues, thereby enhancing the overall accuracy and relevance of the generated answers.

\begin{table*}[htbp]
  \centering
      \caption{
  \label{tab:prompt}
    \textcolor{black}{
    The results of models (using Qwen2-VL (2B)) with different ways to prompt the LLM for generating the answer.
    ``Vision Only'' and ``Text Only'' mean only the A-GCN output for the CT slices and question tokens are used as the soft prompt to instruct the LLM to produce the answer, respectively.
    ``LLM'' and ``Ours'' are reference models standing for the vanilla LLM baseline and our full model, respectively.
    }
    }
  \begin{tabular}{l | c | c c c c c | c}
    \toprule
    Approach & Metric & Plane & Phase & Organ & Abnormality & Location & Mean \\
    \midrule
    \multirow{4}{*}{LLM} 
      & BLEU       & 97.87 & 74.02 & 33.82 & 15.46 & 23.26 & 48.89 \\
      & ROUGE      & 97.90 & 78.25 & 37.59 & 18.79 & 27.30 & 51.97 \\
      & METEOR     & 48.73 & 63.41 & 23.30 & 12.45 & 18.09 & 33.20 \\
      & BERT-Score & 98.94 & 95.16 & 88.53 & 85.71 & 87.13 & 91.09 \\
    \midrule
    \multirow{4}{*}{Ours (Vision Only)} 
      & BLEU       & 98.28 & 74.37 & 34.24 & 15.98 & 24.01 & 49.38 \\
      & ROUGE      & 98.24 & 78.46 & 37.97 & 19.07 & 27.62 & 52.27 \\
      & METEOR     & 49.27 & 63.59 & 23.65 & 12.58 & 18.50 & 33.52 \\
      & BERT-Score & 99.33 & 95.40 & 89.01 & 86.04 & 87.47 & 91.45 \\
    \midrule
    \multirow{4}{*}{Ours (Text Only)} 
      & BLEU       & 98.24 & 74.40 & 34.29 & 16.02 & 23.97 & 49.38 \\
      & ROUGE      & 98.20 & 78.41 & 37.93 & 19.10 & 27.59 & 52.25 \\
      & METEOR     & 49.22 & 63.60 & 23.60 & 12.55 & 18.47 & 33.49 \\
      & BERT-Score & 99.20 & 95.37 & 89.04 & 85.99 & 87.43 & 91.41 \\
        \midrule
    \multirow{4}{*}{Ours} 
      & BLEU       & \textbf{98.57} & \textbf{74.72} & \textbf{34.58} & \textbf{16.33} & \textbf{24.43} & \textbf{49.73} \\
      & ROUGE      & \textbf{98.52} & \textbf{78.81} & \textbf{38.39} & \textbf{19.52} & \textbf{28.02} & \textbf{52.65} \\
      & METEOR     & \textbf{49.51} & \textbf{63.91} & \textbf{24.03} & \textbf{13.18} & \textbf{18.84} & \textbf{33.89} \\
      & BERT-Score & \textbf{99.62} & \textbf{95.73} & \textbf{89.45} & \textbf{86.52} & \textbf{87.72} & \textbf{91.81} \\
    \bottomrule
  \end{tabular}
\end{table*}

\textcolor{black}{
We compare our approach with existing studies and report the results in Table \ref{tab:sota}.
Our approach demonstrates superior or comparable performance to existing approaches across multiple evaluation metrics. 
Specifically, our model consistently outperforms the RadFM baseline and achieves competitive results relative to the M3D model. 
Although the M3D model benefits from training on a larger dataset, our approach leverages a cross-modal feature graph along with attentive graph convolutional networks (A-GCN) to effectively capture cross-slice and cross-modal relationships. 
This enables our model to better aggregate critical visual and textual cues, particularly in challenging aspects such as abnormality detection and precise localization. 
}

\subsection{Effect of Different LLMs}

We evaluated the impact of the type and scale of LLMs, where we utilize Qwen2-VL (7B) and LLaVA-med (7B) as the LLM.
The results of the baseline and our approach are reported in Table \ref{tab:llm-types}.
As presented in the Table, our approach with different LLMs outperforms the vanilla LLM on various evaluation metrics, which further confirms the effectiveness and robustness of our approach.
Moreover, compared to the results of our approach using Qwen2-VL (2B) in Table \ref{tab:overall}, the results of using a larger LLM in the same type (i.e., the Qwen2-VL (7B)) are higher, which is explained by larger models have better capabilities for capturing the information in the input and produce superior answers. 
Meanwhile, the LLM (i.e., LLaVA-med) trained on medical data allows the model to perform better, which is intuitive since the LLM has already learn medical domain information and is able to understand medical data better.

\subsection{Effect of Training Strategies}

To test the effect of different training strategies on the model performance, we try full parameter fine-tuning (which is the default setting) and a representative parameter efficient fine-tuning (PEFT) approach named LoRA when using Qwen2-VL (2B) as the LLM.
The results of using full parameter fine-tuning and LoRA are reported in Table \ref{tab:peft}.
The results show that our approach works with different training strategies, where our approach outperforms the baseline model under different settings.

\begin{figure*}[t]
    \centering
    \includegraphics[width=1\textwidth,trim=0 10 0 0]{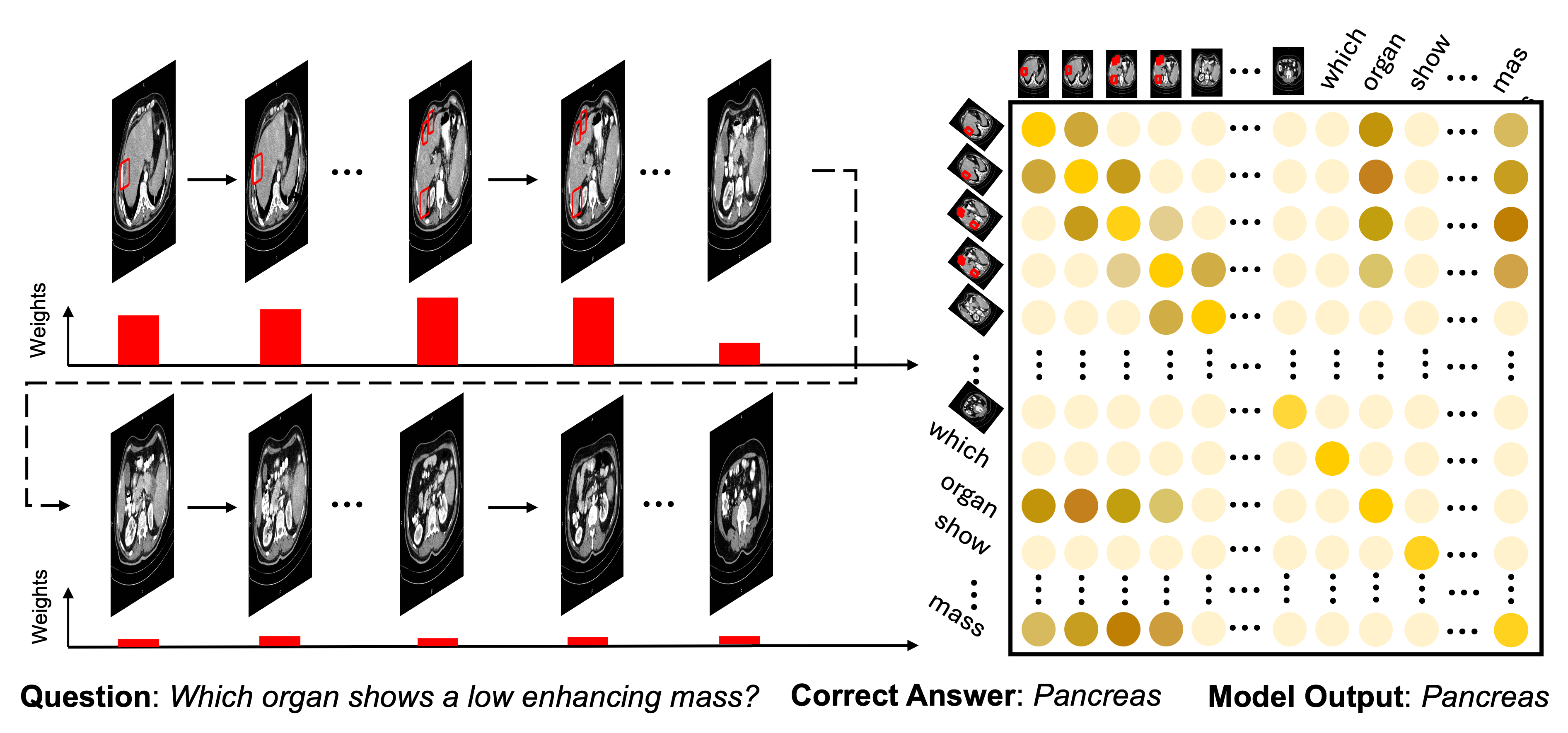}
    \caption{A case study for CT VQA. The left part presents the attention weights (in red bars) for different CT slices, where higher bars stand for heavier weights.
    The right side presents the attention across CT slices and question tokens of the last layer of the graph, where deeper color stands for higher weights.
    For better illustration, we highlighted the regions in the CT slices that provide essential hints for answering the question in the red boxes.}
    \label{fig:case_study}
    \vspace{-1.0em}
\end{figure*}

\subsection{Different Prompting Strategies}

Table \ref{tab:prompt} summarizes the performance of our model with Qwen2-VL (2B) under various prompting strategies derived from the A-GCN outputs. 
We compare two settings, where \textbf{Vision Only} uses only slice (visual) representations and \textbf{Text Only} uses only token (text) representations.
We also present the performance of two reference models, namely, the LLM baseline and our full model that concatenates both visual and textual representations.
First, the LLM baseline yields the lowest scores across all evaluation metrics, underscoring the need for explicit multimodal guidance. Without information from our A-GCN, the baseline lacks supplementary context, which limits its ability to integrate critical anatomical and semantic cues necessary for precise CT VQA.
Second, the ``Vision Only'' and ``Text Only'' settings each improve performance compared to the LLM baseline. The ``Vision Only'' approach benefits from the rich anatomical context provided by CT slices but misses the finer semantic details conveyed by the clinical questions. 
The ``Text Only'' strategy captures the nuances in the queries yet fails to incorporate the essential visual information required for localizing and interpreting anatomical structures.
Third, our approach, which fuses the slice and token embeddings, achieves the best overall performance. 
Our approach leverages complementary information from both modalities, allowing the model to effectively reconcile visual details with semantic context and thus lead to improved answer generation in CT VQA tasks.

\subsection{Case Study}

Figure \ref{fig:case_study} illustrates a representative example from the test set with the input CT slices, the question, the gold standard answer, as well as the output from our approach (using Qwen2-VL (2B)), with the red bars illustrating the attention weights assigned to different CT slices, where higher ones refer to heavier weights.
In this case, our approach assigns higher attention weights to a few critical slices that display the key anatomical and pathological features (marked by red boxes) relevant to the clinical question.
By analyzing the attention distributions, we can trace back the slices that contribute most to the generated answer.
The model's ability to explicitly identify and focus on these important slices not only enhances its diagnostic accuracy but also substantially improves interpretability. 
One is able to inspect the attention maps to understand which parts of the CT volume are essential to the answer, thus guiding the trace-back of the evidence towards pathology understanding, thereby increasing the transparency and trustworthiness of the system.

\section{Related Work}

Early medical VQA systems generally follow the paradigm of general-domain VQA,
which conventionally utilize a visual encoder and a textual encoder to process the image and the question, respectively, and then merge these features to produce the answer either by classification (for closed-ended questions) or generation (for open-ended questions) \cite{Antol2015,Abacha2019,li2020oscar, chen2020counterfactual, ding2022mukea, gupta2022swapmix, cho2023generative}. 
To improve performance, many studies leverage advanced architectures or learning processes, such as attention mechanisms, to encode vision and text features \cite{chen2020counterfactual,rahman2021improved,ding2022mukea,cascante2022simvqa,lin2022revive,tian2024diffusion,tian2025recurrent}. 
Particularly for the medical domain, given the limited annotated data for healthcare, researchers further leverage transfer learning and multi-modal pretraining to boost performance \cite{tian2019chimed,AlSadi2021,gong2021cross,li2023masked,zhang2023pmc,van2023open,tian-etal-2024-chimed,chen2024miss}.
Common strategies include using pre-trained backbones for image feature extraction \cite{Antol2015}, employing self-supervised contrastive learning on medical images and text \cite{Lin2023}, and even incorporating meta-learning or multi-task learning (e.g., by adding localization or report generation tasks) to improve generalization \cite{Lin2023}. 
Moreover, several studies integrate domain knowledge through medical ontologies or knowledge graphs to enrich the model’s understanding of clinical concepts \cite{Liu2018,liu2021slake, huang2023medical, naseem2023k, hu2023expert, hu2024interpretable}.
Particularly for CT VQA, existing studies generally apply the standard medical VQA approach to CT volumes, where there are studies that utilize multi-task learning to train a model on CT report generation, CT image segmentation, and CT VQA so as to leverage supervision from multiple tasks to improve CT VQA \cite{bai2024m3d}.
Compared with these studies, our approach explicitly models the inter-slice relationships in CT volumes in fusing visual features with the help of textual evidence from questions, so that captures the continuity and spatial dependencies across slices and enables more nuanced reasoning compared to conventional approaches.

\section{Conclusion}

In this paper, we propose a novel approach for CT VQA by integrating a cross-modal feature graph into an LLM.
Our approach encodes CT slices and clinical queries, constructs a tailored context graph, and employs an attentive GCN to fuse spatial and semantic information, which is then used to prompt the LLM for answer generation.
Experimental results on the M3D-VQA benchmark reveal that our model achieves state-of-the-art performance and exhibits enhanced interpretability by explicitly highlighting key anatomical and clinical cues.
Overall, the cross-modal feature graphing for CT VQA represents a step towards knowledge-aware and interpretable medical multimodal processing, which might inspire further research on integrating domain knowledge for healthcare automation.

% \ifCLASSOPTIONcaptionsoff
%   \newpage
% \fi

\bibliographystyle{IEEEtranS}
\bibliography{reference}

% \newpage

% \appendices

\end{document}